\RequirePackage{fix-cm}
\documentclass[smallextended]{svjour3}       
\smartqed  
\usepackage{graphicx}
\usepackage{amsmath}
\usepackage{amsfonts}
\usepackage{multirow}
\usepackage{color}
\usepackage{soul}
\usepackage[table]{xcolor}
\usepackage[export]{adjustbox}
\usepackage{subfig}
\newcommand{\ccg}{\cellcolor{green!75}}
\newcommand{\ccy}{\cellcolor{yellow!75}}
\newcommand{\ccr}{\cellcolor{red!75}}

\DeclareMathOperator*{\argmax}{argmax}

\begin{document}
\setlength{\abovedisplayskip}{3pt}
\setlength{\belowdisplayskip}{3pt}
\setlength\tabcolsep{2pt}
\title{Active Learning using Deep Bayesian Networks for Surgical Workflow Analysis}


\author{Sebastian Bodenstedt \and Dominik Rivoir \and Alexander Jenke \and Martin Wagner \and Michael Breucha \and Beat M{\"u}ller-Stich \and S{\"o}ren Torge Mees \and J{\"u}rgen Weitz \and Stefanie Speidel}


\institute{S. Bodenstedt \and D. Rivoir \and A. Jenke \and S. Speidel  \at
              Department for Translational Surgical Oncology, National Center for Tumor Diseases (NCT), Partner Site Dresden, Dresden, Germany \\
              \email{Firstname.Lastname@nct-dresden.de}           
           \and
           M. Wagner \and B. M{\"u}ller-Stich\at
            Department of General, Visceral and Transplant Surgery, University of Heidelberg, Heidelberg
           \and 
           M. Breucha \and S.T. Mees \and J. Weitz \at
           Department of Visceral, Thoracic and Vascular Surgery, Faculty of Medicine and University Hospital Carl Gustav Carus, TU Dresden, Dresden, Germany
}

\date{Received: date / Accepted: date}

\maketitle

\begin{abstract}
\textit{Purpose}
For many applications in the field of computer-assisted surgery, such as providing the position of a tumor, specifying the most probable tool required next by the surgeon or determining the remaining duration of surgery, methods for surgical workflow analysis are a prerequisite.
Often machine learning based approaches serve as basis for analyzing the surgical workflow.
In general, machine learning algorithms, such as convolutional neural networks (CNN), require large amounts of labeled data.
While data is often available in abundance, many tasks in surgical workflow analysis need annotations by domain experts, making it difficult to obtain a sufficient amount of annotations.

\textit{Methods}
The aim of using active learning to train a machine learning model is to reduce the annotation effort.
Active learning methods determine which unlabeled data points would provide the most information according to some metric, such as prediction uncertainty.
Experts will then be asked to only annotate these data points.
The model is then retrained with the new data and used to select further data for annotation.
Recently, active learning has been applied to CNN by means of Deep Bayesian Networks (DBN).
These networks make it possible to assign uncertainties to predictions.
In this paper, we present a DBN-based active learning approach adapted for image-based surgical workflow analysis task.
Furthermore, by using a recurrent architecture, we extend this network to video-based surgical workflow analysis.
To decide which data points should be labeled next, we explore and compare different metrics for expressing uncertainty.

\textit{Results}
We evaluate these approaches and compare different metrics on the Cholec80 dataset by performing instrument presence detection and surgical phase segmentation.
Here we are able to show that using a DBN-based active learning approach for selecting what data points to annotate next can significantly outperform a baseline based on randomly selecting data points.
In particular, metrics such as entropy and variation ratio perform consistently on the different tasks.

\textit{Conclusion}
We show that using DBN-based active learning strategies makes it possible to selectively annotate data and thereby reducing the required amount of labeled training in surgical workflow related tasks.
\end{abstract}

\section{Introduction}
\label{sec:intro}
The aim of computer-assisted surgery (CAS) is to provide the surgeon with the right type of assistance at the right moment. 
For many applications in CAS, such as providing the position of a tumor, specifying the most probable tool required next by the surgeon or determining the remaining duration of surgery, analyzing the surgical workflow is a prerequisite.
\emph{Surgical workflow analysis} comprises methods for perceiving and understanding surgical processes in the operating room, generally via data collected from sensors or from human input \cite{lalys2014surgical}.
Since laparoscopic surgeries are performed using an endoscopic camera, a video stream is always available during surgery, making it the obvious choice as input sensor data for workflow analysis.

Several methods in the state-of-the-art for video-based surgical workflow analysis utilize convolutional neural networks (CNNs) for interpreting the video stream \cite{aksamentov2017deep,chen2018endo3d,jin2018sv,twinanda2017endonet,zisimopoulos2018deepphase}.
Deep Neural Networks, such as CNNs, have a high number of parameters that have to be determined during training, which requires a large amount of annotated data.
For many tasks in surgical workflow analysis, expert knowledge is often required for labeling data, making it difficult to obtain a sufficient amount of annotations.
Motivated by the fact that data without annotations is often readily available, multiple methods for pretraining CNNs using unlabeled data for solving surgical workflow related tasks have been recently proposed \cite{bodenstedt2017unsupervised,funke2018temporal,ross2018exploiting,yengera2018less}.
These methods generally exploit information inherent in the unlabeled data to solve an auxiliary task related to the actual problem.
Recently crowdsourcing based approaches have been used to successfully create annotations for simple surgical workflow related tasks in laparoscopy, such as tool segmentation \cite{maier2014can,maier2016crowd}, locating point correspondences \cite{maier2015crowdtruth} and for assessing skills \cite{deal2016crowd}.
More complex tasks, such as surgical phase segmentation, require more task-specific background knowledge, which generally only domain experts, such as surgeons, possess.
Often these experts have limited resources for labeling such data, making it difficult to acquire large, annotated data sets.

A system that could instead actively ask for expert labels only on certain examples, e.g. examples with a high uncertainty, would reduce the total annotation effort and make collecting large, annotated datasets for surgical workflow analysis more feasible.
Such a system is called an \emph{active learning} system \cite{cohn1996active}.
During active learning, an initial model is trained using a small amount of labeled data, the \emph{initial training set}.
An \emph{acquisition function} then determines through a metric, such as uncertainty, which data points should be labeled next.
A new model is then trained on the extended training data \cite{gal2017Active}.

Recently, new methods for estimating uncertainties on the predictions of deep neural networks, such as \emph{Deep Bayesian Networks} (DBN), have been developed \cite{gal2016uncertainty}.
Seeing that such estimates can be used for active learning has motivated Gal et al. \cite{gal2017Active} to formulate acquisition functions based on DBNs.

In this paper, we investigate if an active learning system based on DBNs can successfully guide the annotation process for image- and video-based surgical workflow related tasks and thereby reduce the number of required labels.
For this, we first modify the framework proposed in Gal et al. \cite{gal2017Active} for laparoscopic instrument presence detection and phase segmentation.
Namely, our main contributions are the following:
\begin{enumerate}
 \item Propose a solution for multi-label annotations with DBN-based active learning
 \item Propose a recurrent network for DBN-based active learning with videos
 \item Extend the previous network to allow partial annotation of videos
 \item Evaluate and compare the proposed methods using the publicly available Cholec80 dataset \cite{twinanda2017endonet}.
\end{enumerate}

To the best of our knowledge, we are the first to apply DBN-based active learning to annotate data related to surgical workflow.
Furthermore, as far as we are aware, this is the first work that utilizes DBN-based active learning for video annotation.
\section{Methods}
\label{sec:methods}
In this section, we introduce methods for image-based and video-based active learning for surgical workflow analysis tasks.
The basis of our image-based active learning system is a standard CNN that is transformed into a DBN (section \ref{sec:methods:bnn}).
This serves as basis for performing DBN based active learning on single video frames.
To allow active learning on video data, the DBN is further extended into a recurrent DBN (section \ref{sec:methods:rbnn}).
To use the likelihoods of the DBN to select which data points should be labeled next, several different metrics are possible, which are described in section \ref{sec:methods:acq}.
\subsection{Bayesian Network}
\label{sec:methods:bnn}
\begin{figure}[tbh]
    \centering
    \subfloat[]{
      \includegraphics[width=0.16\textwidth,valign=t]{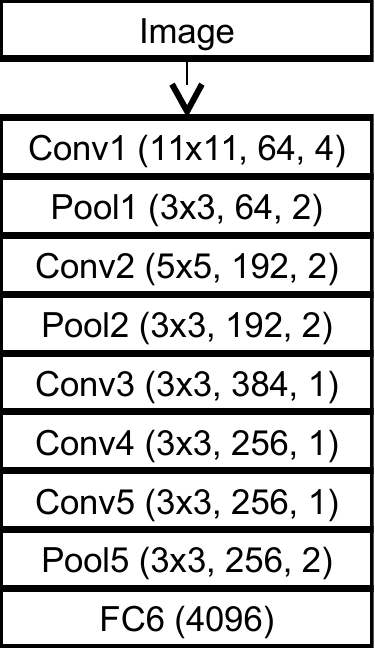}
      \label{fig:bnn:cnn}
    }
    \subfloat[]{
      \includegraphics[width=0.16\textwidth,valign=t]{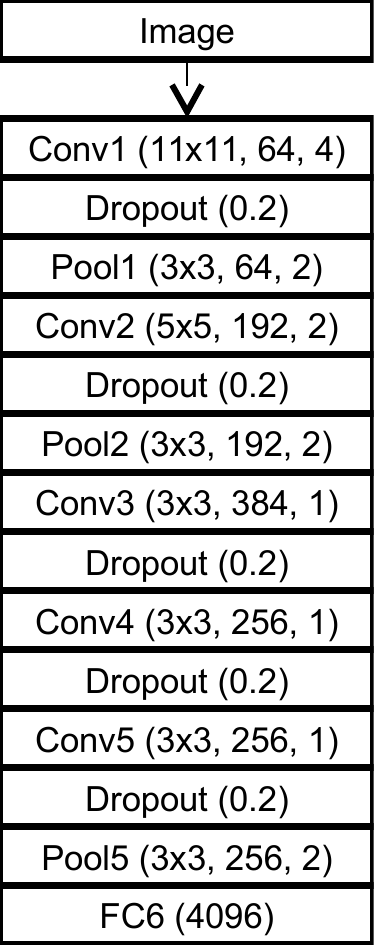}
      \label{fig:bnn:bcnn}
    }
    \subfloat[]{
      \includegraphics[width=0.16\textwidth,valign=t]{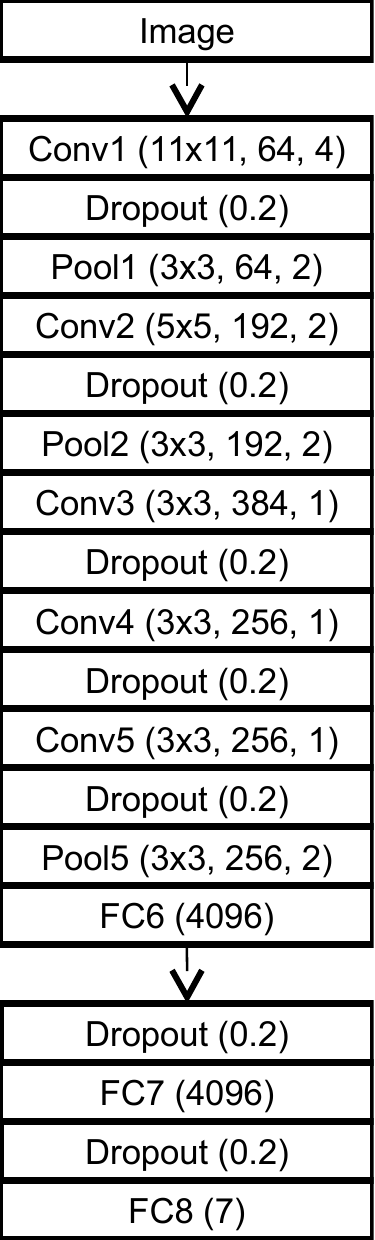}
      \label{fig:apps:cnn}
    }
    \caption{\protect\subref{fig:bnn:cnn} Standard AlexNet without classification layers, \protect\subref{fig:bnn:bcnn} the same AlexNet converted into a DBN through adding dropout layers, and \protect\subref{fig:apps:cnn} the DBN extended for active learning of laparoscopic instrument presence. For convolutional and pooling layers, the numbers indicate filter size, number of feature maps and step size, for fully connected layers, the number of output units and for dropout layers, the probability of setting a value to 0.}
    \label{fig:bnn}
\end{figure}

A standard CNN, based on the AlexNet architecture \cite{Alexnet} and pretrained on ImageNet (see fig. \ref{fig:bnn:cnn}), serves as a foundation of the proposed system for active learning.
We opted to use an AlexNet, as it performed similarly as a ResNet50 during instrument presence detection and phase segmentation, while allowing for faster training.
Active learning requires a method for gauging which unlabeled training examples are ''difficult'' for the current model, e.g. when given an input $x$, an (softmax) output $y$ and training data $D$, determining the likelihood $P(y = l|x,D)$ of label $l$.
While neural networks generally do not output a binary class prediction, but instead a fuzzy prediction, e.g. through a sigmoid or a softmax non-linearity, it has been found that these outputs are not suitable as probability estimates \cite{guo2017calibration}.

DBNs on the other hand are a mathematically proven concept for estimating likelihoods for predictions \cite{gal2016uncertainty}.
DBNs are deep neural networks with a prior probability distribution, such as a Gaussian prior, placed over the weights $W$ of the network: $P(W)$.
The likelihood of a classification is then defined as 
$$P(y = l|x,W) = \text{softmax}(f_W(x))$$
where $f_W(x)$ is the output of the network depend on weights $W$.
Inference in DBNs requires the posterior $P(W|D)$, which is extremely difficult to infer.
Instead, the posterior can be approximated through Monte Carlo dropout, which is done by performing random dropout on every weight layer during training and testing.
Monte Carlo dropout can be shown to be equivalent to performing approximate variational interference, which minimizes the Kullback-Leibler divergence to the true posterior:
$$P(y = l|x,D) = \int P(y = l|x,W) P(W|D) dW \approx \frac{1}{T}\sum_{t=1}^T P(y = l|x,\hat{W}_t)$$
with $\hat{W}_t \sim q_\theta(W)$, where $q_\theta(W)$ is the dropout distribution \cite{gal2016uncertainty}.
In other words, to determining the likelihood of a classification of a sample $x$ during testing, we classify the sample $T$ times using Monte Carlo dropout and average the outputs of the softmax.

The previous CNN that has been extended into a DBN can be seen in fig. \ref{fig:bnn:bcnn}.
By applying task-specific classification layers to the network, predictions and their likelihood can be estimated.

\subsection{Recurrent Bayesian Network}
\label{sec:methods:rbnn}
Many tasks in surgical workflow analysis, such as phase segmentation, require that frames are viewed in the context of an entire video or at least in the context of previous frames.
Recurrent neural networks (RNN) make such an analysis possible by introducing recurrence into the topology of a network.
This allows information from previous frames to contribute to future predictions.

Long short-term memory units (LSTM), a more complex form of the RNN, can learn to strategically remember, but also forget, information from previously seen inputs, while forgoing the problem of vanishing gradients common to RNNs \cite{hochreiter1997long}.
Combining CNNs with LSTMs makes video-based workflow analysis, by using exclusively deep neural networks, possible \cite{bodenstedt2017unsupervised,chen2018endo3d,funke2018temporal,yengera2018less}.

\begin{figure}[tb]
    \centering
    \subfloat[]{
      \includegraphics[width=0.16\textwidth,valign=t]{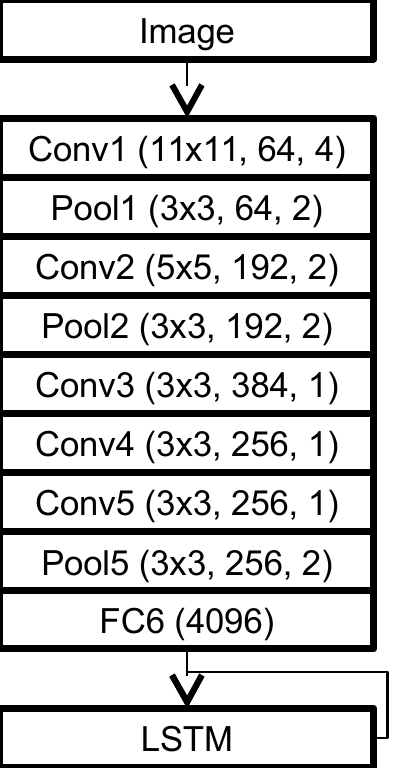}
      \label{fig:brnn:crnn}
    }
    \subfloat[]{
      \includegraphics[width=0.32\textwidth,valign=t]{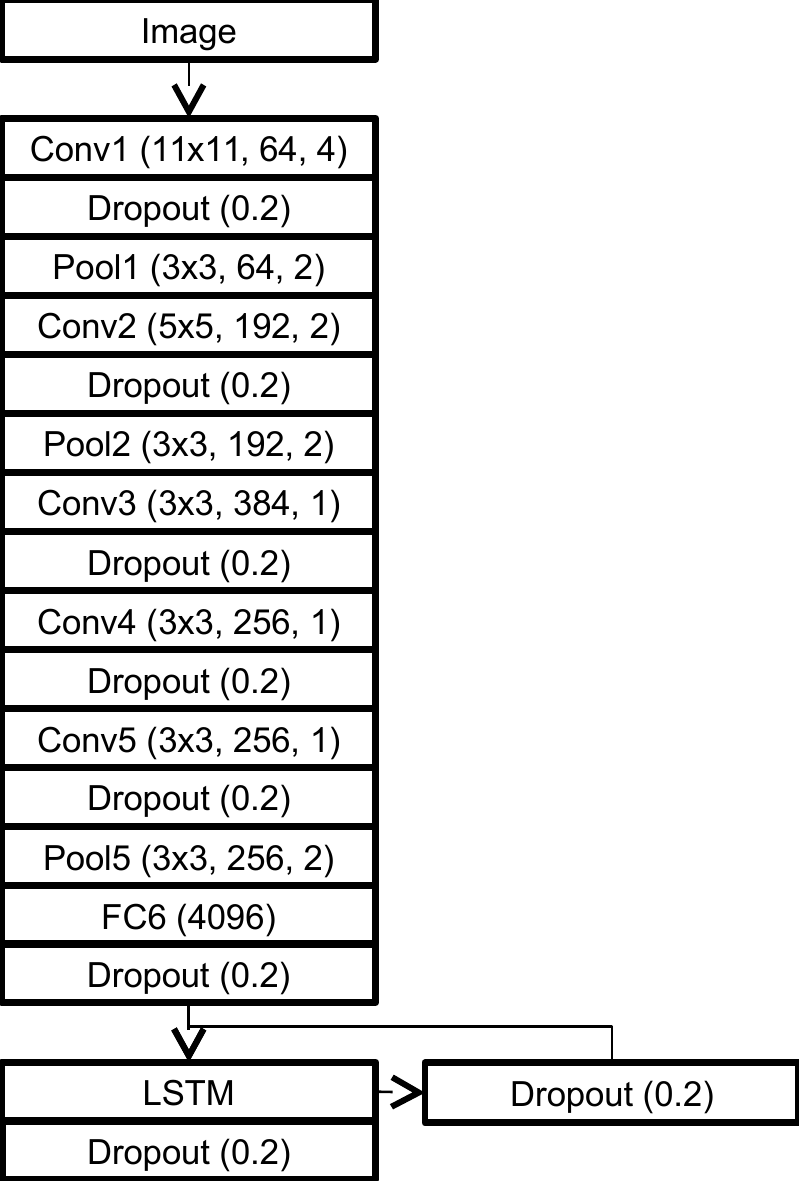}
      \label{fig:brnn:brcnn}
    }
    \subfloat[]{
      \includegraphics[width=0.32\textwidth,valign=t]{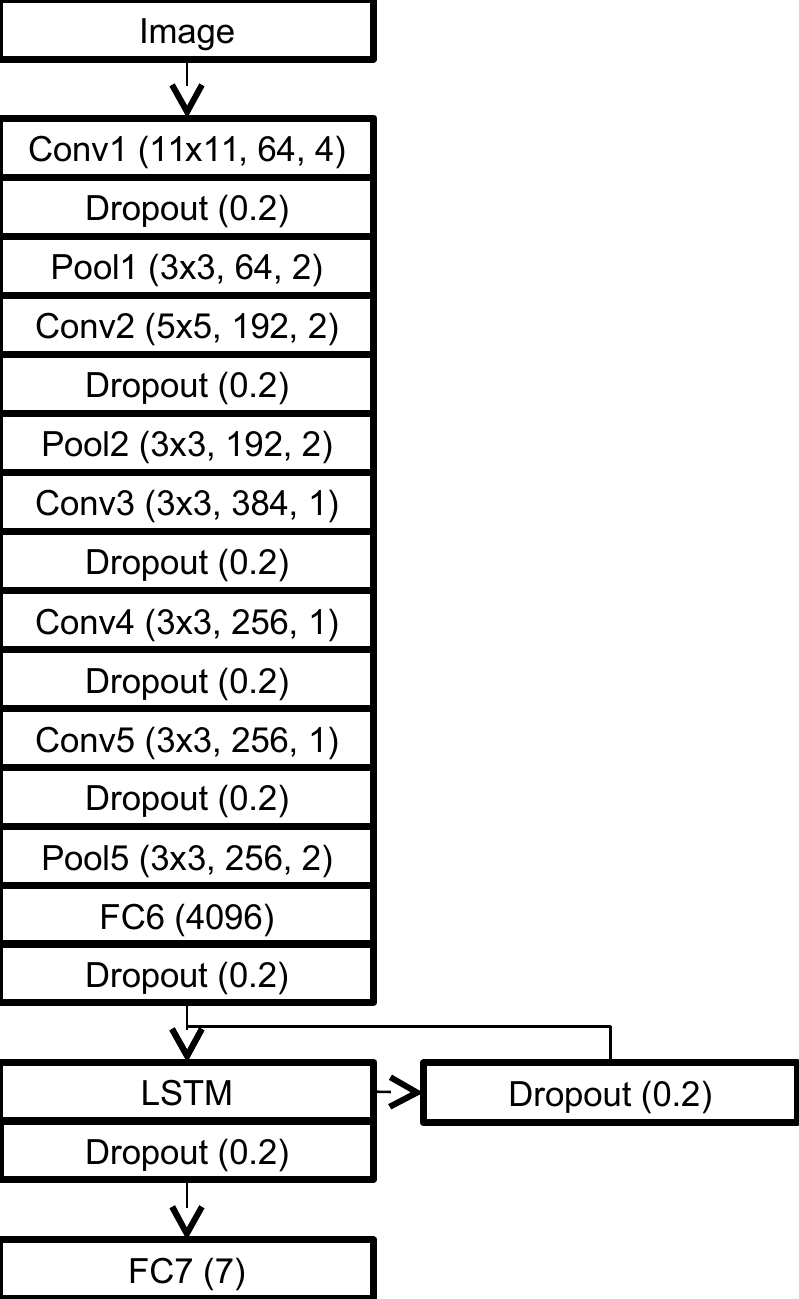}
      \label{fig:apps:rcnn}
    }
    \caption{\protect\subref{fig:brnn:crnn} AlexNet without classifier extended with an LSTM, \protect\subref{fig:brnn:brcnn} the same network converted into a recurrent DBN through adding dropout layers, and \protect\subref{fig:apps:rcnn} the DBN extended for surgical phase segmentation. For convolutional and pooling layers, the numbers indicate filter size, number of feature maps and step size, for fully connected layers and LSTMs, the number of output units and for dropout layers the probability of setting a value to 0.}
    \label{fig:brnn}
\end{figure}

\begin{figure}[tb]
    \centering
      \includegraphics[width=0.7\textwidth]{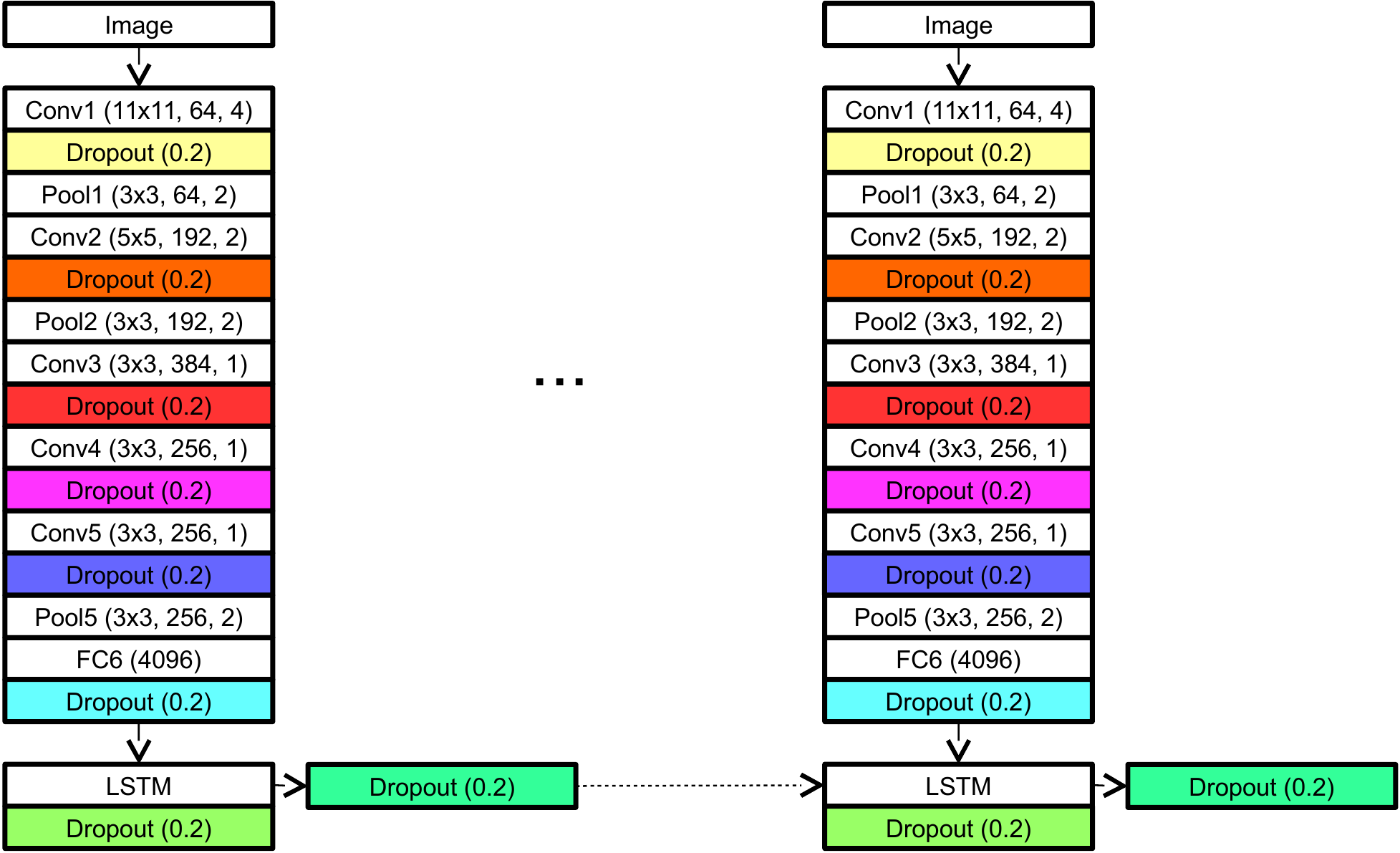}
    \caption{Modified recurrent DBN in multiple steps of a sequence. The colors indicate identical dropout masks.}
    \label{fig:mbrnn}
\end{figure}

By applying the paradigm described in section \ref{sec:methods:bnn}, we can extend the topology of a CNN-LSTM based on AlexNet \cite{Alexnet} (see fig. \ref{fig:brnn:crnn}) into that of a Bayesian CNN-LSTM (see. fig. \ref{fig:brnn:brcnn}). 
One approach to perform inference with this network would be to naively apply random dropouts independently to each weight layer for every element in a given sequence.
Multiple studies though indicate that such a naive dropout has negative effects on RNNs, such as added noise and a disruption of dynamics \cite{gal2016theoretically}.
As an alternative, the authors in \cite{gal2016theoretically} propose a theoretically grounded variant of dropout for LSTMs.
The idea is to sample dropout masks for each layer in the recurrent DBN at the beginning of each sequence and to use the same mask for each time-step (see fig. \ref{fig:mbrnn}).
The naive approach would be equivalent to sampling new masks at every time-step.

This recurrent DBN makes video-based classification possible, while simultaneously allowing likelihood estimations for each classification.
\subsection{Acquisition Functions}
\label{sec:methods:acq}
Given a DBN with weights $W$ and a pool with unlabeled data points $\bar{D}$, the active learning framework uses an acquisition function $a(f_W(x))$, with $x \in \bar{D}$, to determine which data points show high levels of uncertainty.
The following criteria is used to select which data points should be labeled next:
$$x^* = \argmax_{x \in \bar{D}} a(f_W(x)).$$

The authors in \cite{gal2017Active} propose multiple acquisition functions that have to be evaluated for their suitability in active learning for surgical workflow tasks.
\paragraph{Variance}
One simple metric for measuring the uncertainty is to compute the variance of the different likelihoods contributing to the posterior:
$$\text{Var}(x) = \mathbb{E}[(P(y = l|x,\hat{W}_t) - \mu)^2]$$
with $\mu = \mathbb{E}[P(y = l|x,\hat{W}_t)]$.
Variance measures how the $T$ likelihood predictions are spread around their arithmetic mean.
Here we assume that a large spread corresponds to a large amount of uncertainty.
\paragraph{Variation Ratio (VR)}
Similarly to variance, the variation ratios also measures the spread of the $T$ predictions, in this case around the mode, i.e. the most common predicted class.
$$\text{VR}(x) = 1 - \frac{f_m}{T}$$
where $f_m$ is the frequency of the mode in the $T$ predictions.
\paragraph{Entropy}
A further possibility for measuring the uncertainty of the posterior likelihood is using predictive entropy from information theory:
$$\mathbb{H}(x, \bar{D}) = - \sum_l  P(y = l|x,\bar{D})\log P(y = l|x,\bar{D})$$
$\mathbb{H}$ reaches its maximum when the likelihood of all classes becomes equal. Its minimum (zero) is reached when the likelihood of a single class is equal to one.
\paragraph{Mutual Information (MI)}
An extension of predictive entropy is to examine the mutual information between the posterior and the likelihoods of the $T$ predictions:
$$\mathbb{I}(x) = \mathbb{H}(x, \bar{D}) - \mathbb{E}(H(x,\hat{W}_t))$$
\section{Applications}
To evaluate the suitability of the DBNs described in the previous section for active learning in workflow analysis tasks, we examine two different applications.
In section \ref{sec:methods:ins} we extended a DBN to perform active learning for laparoscopic instrument presence detection and in section \ref{sec:methods:phase} a recurrent DBN is used to perform active learning for surgical phase segmentation.
For both tasks, the publicly available Cholec80 dataset \cite{twinanda2017endonet} is used. 
It consists of 80 videos from laparoscopic cholecystectomies, in which surgical instrument presence and surgical phases have been annotated.
\subsection{Instrument Presence}
\label{sec:methods:ins}
To perform active learning for instrument presence detection, we extend the DBN proposed in section \ref{sec:methods:bnn} with two further fully connected layers (see fig. \ref{fig:apps:cnn}).
The last layer consists out of 7 units, one for each instrument type in Cholec80.
Since multiple instruments of different types can be visible in the same video frame, a sigmoid nonlinearity is used on the final layer instead of a softmax.
During training, we use a weighted DICE-loss \cite{milletari2016v} as cost function.

As this is a multi-label problem, when computing the uncertainty of the prediction of a given image using, any one of the acquisition functions $a(f_W(x))$ outlined in section \ref{sec:methods:acq} will not return a scalar value, but instead a 7-dimensional vector containing the uncertainty of each class.
To reduce this vector to a scalar, we examine the suitability of two different methods for aggregation: $$a_{\text{mean}} = \mathbb{E}(a(f_W(x)))\text{ and }a_{\text{max}} = \max(a(f_W(x))).$$
The frames with the highest uncertainty from $\bar{D}$ are then selected for annotation.
$a_{\text{mean}}$ would here favor frames with a high certainty across all classes, while $a_{\text{max}}$ would favor frames in which one class shows a large amount uncertainty, regardless of the uncertainties of the other classes.
\subsection{Phase Segmentation}
\label{sec:methods:phase}
To allow active learning for phase segmentation, we extend the recurrent DBN proposed in section \ref{sec:methods:rbnn} by adding a fully connected output layer (see fig. \ref{fig:apps:rcnn}).
The layer has 7 output units, one for each surgical phase in Cholec80, with a softmax nonlinearity.
During training, we use cross-entropy as cost function.

We assume that judging the current surgical phase from a single video frame is difficult and prone to ambiguities, we therefore opt to query for annotation for temporally connect segments.
For this, we propose two methods for selecting the next queries from $\bar{D}$.
\paragraph{Video-based}
A naive approach would be to determine which unlabeled video from $\bar{D}$ has the largest amount of uncertainty and ask an expert to annotate this video completely.
Given a video, we compute the uncertainty for each frame and aggregate these uncertainties using either $a_{\text{max}}$ or $a_{\text{mean}}$, where $a_{\text{mean}}$ would favor videos with a high overall uncertainty and $a_{\text{max}}$ would tend to select videos where a single frame exhibits a high uncertainty.

\paragraph{Segment-based}
Annotating an entire video is a time-consuming process, which is also difficult to parallelize.
Instead it would be preferable to select the most uncertain parts of a video and query an expert to just annotate these.

To accomplish this, we divide each video into segments with a length of 5 minutes.
During active learning, we determine the uncertainty of each segment in the same manner as described above for an entire video.
We then query for the most uncertain segments according to either $a_{\text{max}}$ or $a_{\text{mean}}$.

This leads to having incompletely labeled videos during training, which is a problem as the recurrent nature of the network requires that each sequence be trained from the beginning.
To account for this, we slightly modify the cost function. 
Given the output $y_i$ and the correct label $l_i$ of frame $i$ and the cost $C(y_i, l_i)$, we know define $\hat{C}(y_i, l_i) = m_i \cdot C(y_i, l_i)$ with $m_i \in \lbrace 0, 1 \rbrace$, depending on whether $i$ is annotated or not.
This causes frames whose label is unknown at this point, to be excluded from the overall cost, while still preserving their influence on the predictions of the annotated frames.
\begin{table}[bt]
\centering
\resizebox{\textwidth}{!}{
\begin{tabular}{l|c|c|c|c|c|c|c|c|c}
\% data&\multicolumn{2}{c|}{Variance}&\multicolumn{2}{c|}{VR}&\multicolumn{2}{c|}{Entropy}&\multicolumn{2}{c|}{MI}&Random\\
annotated	&Max			&Mean				&Max			&Mean			&Max			&Mean				&Max&Mean&\\
\hline
20\%		&\ccr78\% (92\%)		&\ccg\textbf{79\% (92\%)}		&\ccy\textbf{79\% (92\%)}	&\ccg\textbf{79\% (92\%)}	&\ccr78\% (92\%)		&\ccy\textbf{79\% (92\%)}		&\ccr78\% (92\%)		&\ccy\textbf{79\% (92\%)}	&78\% (92\%)\\
30\%		&\ccg81\% (93\%)		&\ccg\textbf{82\% (93\%)}		&\ccr80\% (92\%)		&\ccg\textbf{82\% (93\%)}	&\ccy81\% (93\%)		&\ccg\textbf{82\% (93\%)}		&\ccg79\% (93\%)		&\ccr80\% (93\%)		&79\% (92\%)\\
40\%		&\ccg82\% (93\%)		&\ccg\textbf{83\% (94\%)}		&\ccr81\% (93\%)		&\ccg82\% (93\%)		&\ccr81\% (93\%)		&\ccg82\% (94\%)			&\ccy80\% (93\%)		&\ccr80\% (93\%)		&81\% (93\%)\\
50\%		&\ccg\textbf{83\% (94\%)}	&\ccg\textbf{83\% (94\%)}		&\ccr82\% (93\%)		&\ccg83\% (93\%)		&\ccr82\% (93\%)		&\ccr82\% (94\%)			&\ccr82\% (93\%)		&\ccr82\% (93\%)		&82\% (93\%)\\
60\%		&\ccg\textbf{83\% (94\%)}	&\ccg\textbf{83\% (94\%)}		&\ccr82\% (93\%)		&\ccg83\% (93\%)		&\ccr82\% (94\%)		&\ccy82\% (94\%)			&\ccy82\% (94\%)		&\ccy82\% (93\%)		&82\% (93\%)\\
\end{tabular}}
\caption{Test performance on the instrument presence detection task. Each line shows how the error progresses when more data is selected with one the acquisition functions, in combination with $a_{\text{max}}$ and $a_{\text{mean}}$, or the baseline. The values shown are the weighted F1-score and, in parenthesis, accuracy. The best performances for each line are in bold. The statistical significance is color-coded: Green indicates $p < 0.01$, yellow $p < 0.05$ and red $p \ge 0.05$.}
\label{tab:res_ins}
\end{table}
\section{Evaluation}
\label{sec:eval}
As previously stated, we evaluate our proposed active learning methods for surgical workflow analysis tasks (instrument presence and phase segmentation) on the Cholec80 dataset.
For this, we first divide the dataset in 4 subsets of 20 videos each, as outlined in \cite{funke2018temporal}.
Each video was sampled at a rate of one frame per second and each frame was downsampled to a resolution of $384 \times 216$ pixels.
No methods for data augmentation were applied.

During evaluation, we proceed in an identical manner for both instrument presence detection and phase segmentation.
We begin by dividing the 4 subsets into a training data set (subsets 1-3) and testing data set (subset 4).
We select the first 6 videos (10\%) from the training data set and define the remaining 54 video as $\bar{D}$.
The 6 videos and their annotations are used to train an initial DBN using the Adam optimizer \cite{kingma2014adam}.
To ensure repeatability and comparability, the layers which have not been pretrained, are initialized with identical values for each experiment.
We train for 100 epochs or until the training cost reaches $5\cdot10^{-4}$.
After training, we note the performance on the test data, namely the weighted F1-score for each class label and accuracy, and proceed to select data points from $\bar{D}$ using one of the acquisition functions in \ref{sec:methods:acq} and aggregate using either $a_{\text{max}}$ or $a_{\text{mean}}$.
New data points are then selected until a further 10\% of the training data set has been annotated. We then train the model again from scratch using all the available annotated data, noting the performance on the test data after each training run is completed.

As baseline, we use a fifth acquisition function that selects data points at random.
For each task, the baseline is computed 4 times and the results are averaged.
Given the averaged baseline and results from one of the introduced acquisition functions, we perform a Wilcoxon signed-rank test to assess statistical significance of the performance changes in the F1-score \cite{Wilcoxon45}.

\paragraph{Instrument Presence}
For the instrument presence task, the DBN was trained with a learning rate of $10^{-6}$, a L2-norm based weight decay of $10^{-4}$ and a batch size of 128.
Data points in $\bar{D}$ for this task were essentially every frame in the training data, meaning we did not incorporate any knowledge about the structure of the original videos while querying for new frames.
We opted to only display results up to 60\% as this shows the most drastic differences, most of the methods, including the baseline converged to a F1-score of 83\% and an accuracy of 94\% shortly after.
The results of the active learning process with the different acquisition functions in comparison to the baseline can be found in table \ref{tab:res_ins}.

\paragraph{Phase Segmentation}
For the instrument presence task, the recurrent DBN was trained with a learning rate of $5\cdot10^{-5}$, a L2-norm based weight decay of $10^{-4}$ and a batch size of 128.
Similar to the instrument presence task, we opted to display results up to 60\% as this shows the most drastic differences, most of the methods, including the baseline converged to a F1-score of 86\% and an accuracy of 92\% shortly after.
The results for the video-based phase segmentation can be seen in table \ref{tab:res_phase_comp} and the results for the segment-based video segmentation in table \ref{tab:res_phase_seg}.

\section{Discussion}
\label{sec:discussion}
\begin{table}[bt]
\centering
\subfloat[Video-based]{
\label{tab:res_phase_comp}
\resizebox{\textwidth}{!}{
\begin{tabular}{l|c|c|c|c|c|c|c|c|c}
\% data&\multicolumn{2}{c|}{Variance}&\multicolumn{2}{c|}{VR}&\multicolumn{2}{c|}{Entropy}&\multicolumn{2}{c|}{MI}&Random\\
annotated&Max&Mean&Max&Mean&Max&Mean&Max&Mean&\\
\hline
20\%			&\ccr66\% (77\%)			&\ccg68\% (77\%)			&\ccr67\% (79\%)			&\ccy68\% (76\%)			&\ccg\textbf{71\% (80\%)}		&\ccr66\% (78\%)			&\ccy67\% (81\%)			&\ccr65\% (73\%)			&64\% (76\%)\\
30\%			&\ccr68\% (78\%)			&\ccg\textbf{76\% (85\%)}		&\ccr69\% (81\%)			&\ccg73\% (82\%)			&\ccg72\% (83\%)			&\ccy71\% (80\%)			&\ccg75\% (84\%)			&\ccy70\% (81\%)			&67\% (79\%)\\
40\%			&\ccr73\% (80\%)			&\ccg\textbf{79\% (88\%)}		&\ccr74\% (84\%)			&\ccg79\% (87\%)			&\ccr75\% (82\%)			&\ccr75\% (86\%)			&\ccy76\% (83\%)			&\ccg78\% (87\%)			&74\% (84\%)\\
50\%			&\ccr77\% (87\%)			&\ccr78\% (86\%)			&\ccg81\% (90\%)			&\ccg\textbf{82\% (90\%)}		&\ccr77\% (85\%)			&\ccg80\% (88\%)			&\ccr77\% (85\%)			&\ccr78\% (88\%)			&77\% (84\%)\\
60\%			&\ccr80\% (87\%)			&\ccr80\% (87\%)			&\ccy81\% (90\%)			&\ccg\textbf{82\% (91\%)}		&\ccr79\% (89\%)			&\ccy80\% (89\%)			&\ccr80\% (88\%)			&\ccr80\% (90\%)			&80\% (86\%)\\
\end{tabular}}
}

\subfloat[Segment-based]{
\label{tab:res_phase_seg}
\resizebox{\textwidth}{!}{
\begin{tabular}{l|c|c|c|c|c|c|c|c|c}
\% data&\multicolumn{2}{c|}{Variance}&\multicolumn{2}{c|}{VR}&\multicolumn{2}{c|}{Entropy}&\multicolumn{2}{c|}{MI}&Random\\
annotated&Max&Mean&Max&Mean&Max&Mean&Max&Mean&\\
\hline
20\%			&\ccr62\% (78\%)			&\ccr59\% (75\%)			&\ccg\textbf{71\% (82\%)}		&\ccr64\% (76\%)			&\ccg68\% (79\%)			&\ccg67\% (74\%)			&\ccr63\% (74\%)			&\ccg71\% (78\%)			&62\% (75\%)\\
30\%			&\ccr70\% (83\%)			&\ccr68\% (79\%)			&\ccg76\% (85\%)			&\ccr74\% (85\%)			&\ccg\textbf{79\% (87\%)}		&\ccy74\% (85\%)			&\ccr73\% (87\%)			&\ccr73\% (85\%)			&73\% (84\%)\\
40\%			&\ccr75\% (89\%)			&\ccr75\% (87\%)			&\ccg79\% (87\%)			&\ccg\textbf{81\% (88\%)}		&\ccg80\% (88\%)			&\ccg79\% (87\%)			&\ccr72\% (80\%)			&\ccg79\% (88\%)			&76\% (86\%)\\
50\%			&\ccy80\% (89\%)			&\ccr77\% (86\%)			&\ccg81\% (89\%)			&\ccy81\% (90\%)			&\ccg\textbf{83\% (89\%)}		&\ccg81\% (90\%)			&\ccr78\% (86\%)			&\ccy81\% (88\%)			&79\% (88\%)\\
60\%			&\ccg81\% (91\%)			&\ccg84\% (91\%)			&\ccg\textbf{85\% (91\%)}		&\ccy80\% (91\%)			&\ccg83\% (90\%)			&\ccg83\% (92\%)			&\ccg81\% (91\%)			&\ccg84\% (92\%)			&79\% (89\%)\\
\end{tabular}}
}
\caption{Test performance on the phase segmentation task using completely annotated videos \protect\subref{tab:res_phase_comp} and annotated video segments \protect\subref{tab:res_phase_seg}. Each line shows how the error progresses when more data is selected with one the different acquisition functions, in combination with $a_{\text{max}}$ and $a_{\text{mean}}$, or the baseline. The values shown are the weighted F1-score and accuracy. The best performances for each line are in bold. The statistical significance is color-coded: Green indicates $p < 0.01$, yellow $p < 0.05$ and red $p \ge 0.05$.}
\label{tab:res_phase}
\end{table}
As tables \ref{tab:res_ins}, \ref{tab:res_phase_comp} and \ref{tab:res_phase_seg} clearly show, the DBN-based acquisition functions for active learning generally outperform a baseline based on randomly selecting the next data points.
In the case of the instrument presence task, the methods based on $a_{\text{mean}}$ seem to outperform their counterpart based on $a_{\text{max}}$, indicating that selecting frames on which the uncertainty is spread among multiple classes is the better strategy.
Especially the combination of $a_{\text{mean}}$ and the variance-based acquisition function seems to be the method of choice for this task as it consistently achieves the highest performance.
Furthermore, it can be observed that actively selecting which data points to include next can lead to a disproportionate increase in occurrence of classes that are underrepresented in the data. 
Table \ref{tab:dis:ins} shows that certain instruments classes, for example the bipolar, scissors and clipper, are selected with a higher frequency when compared to a random baseline.
These classes have the lowest rate of occurrence in the dataset.

Similarly, in the phase segmentation task using video-based selection, the methods based on $a_{\text{mean}}$ generally outperform their counterpart based on $a_{\text{max}}$.
The variance-based acquisition function performs also well on this task, though the variation rate-based method performs better, actually outperforming the other methods at 50\% and 60\%.

Interestingly, in the case of the phase segmentation task using segment-based selection, the methods based on $a_{\text{max}}$ seem to be preferable.
This indicates that segments containing large peaks of uncertainty seem to add more information than segments with a more distributed uncertainty. 
Here, the variation ratio and the entropy-based methods seem to perform best.
It can also be noted that the segment-based methods generally produces similar results as the video-based methods with less annotated data, meaning that partially annotating videos seems to be a valid strategy.
Similarly as during the instrument presence task, it can be observed that certain classes are selected with a disproportionate frequency for annotation.
Table \ref{tab:dis:Phase} shows for example that almost all segments containing frames pertaining to P1 are selected for annotation in the first few rounds.
On the other hand, P2 and P4, the longest phases, are selected disproportionately less than with the random baseline.

Overall it can be noted that the acquisition functions based on variation ratio and on entropy, while not always providing the best results, seem to perform consistently well on all tasks, indicating that they might be the best choice when examining a new problem.

\begin{table}[tb]
\centering
\subfloat[Instrument type occurrence]{
\label{tab:dis:ins}
\resizebox{\textwidth}{!}{
\begin{tabular}{l|c|c|c|c|c|c|c||c|c|c|c|c|c|c}
\% anno-&\multicolumn{7}{c||}{Variance + $a_{\text{mean}}$}&\multicolumn{7}{c}{Random}\\
tated&Grasper&Bipolar&Hook&Scissors&Clipper&Irrigator&Bag&Grasper&Bipolar&Hook&Scissors&Clipper&Irrigator&Bag\\
\hline
10\%&60\%(9\%)&5\%(10\%)&51\%(8\%)&3\%(15\%)&3\%(9\%)&9\%(13\%)&5\%(8\%)&60\%(9\%)&5\%(10\%)&51\%(8\%)&3\%(15\%)&3\%(9\%)&9\%(13\%)&5\%(8\%)\\
20\%&57\%(20\%)&10\%(38\%)&47\%(16\%)&3\%(32\%)&4\%(23\%)&10\%(36\%)&7\%(23\%)&56\%(19\%)&5\%(19\%)&54\%(18\%)&2\%(26\%)&3\%(20\%)&7\%(23\%)&6\%(18\%)\\
30\%&56\%(30\%)&10\%(58\%)&42\%(22\%)&4\%(68\%)&7\%(65\%)&11\%(59\%)&8\%(39\%)&56\%(30\%)&5\%(29\%)&54\%(28\%)&2\%(35\%)&3\%(30\%)&6\%(32\%)&6\%(29\%)\\
40\%&55\%(40\%)&8\%(67\%)&44\%(31\%)&4\%(79\%)&6\%(77\%)&10\%(73\%)&7\%(49\%)&55\%(40\%)&5\%(39\%)&55\%(39\%)&2\%(44\%)&3\%(40\%)&6\%(41\%)&6\%(39\%)\\
50\%&54\%(49\%)&7\%(75\%)&46\%(41\%)&3\%(85\%)&5\%(84\%)&9\%(82\%)&7\%(55\%)&55\%(50\%)&5\%(49\%)&55\%(49\%)&2\%(54\%)&3\%(50\%)&6\%(51\%)&6\%(49\%)\\
60\%&53\%(58\%)&6\%(80\%)&49\%(53\%)&3\%(89\%)&5\%(88\%)&8\%(87\%)&6\%(60\%)&55\%(60\%)&5\%(59\%)&55\%(59\%)&2\%(63\%)&3\%(60\%)&6\%(60\%)&6\%(59\%)\\
\end{tabular}}
}

\subfloat[Phase occurrence]{
\label{tab:dis:Phase}
\resizebox{\textwidth}{!}{
\begin{tabular}{l|c|c|c|c|c|c|c||c|c|c|c|c|c|c}
\% anno-&\multicolumn{7}{c||}{Entropy + $a_{\text{max}}$}&\multicolumn{7}{c}{Random}\\
tated&P1&P2&P3&P4&P5&P6&P7&P1&P2&P3&P4&P5&P6&P7\\
\hline
10\%&12\%(21\%)&31\%(6\%)&8\%(9\%)&31\%(9\%)&4\%(7\%)&11\%(12\%)&3\%(8\%)&12\%(21\%)&31\%(6\%)&8\%(9\%)&31\%(9\%)&4\%(7\%)&11\%(12\%)&3\%(8\%)\\
20\%&9\%(37\%)&40\%(20\%)&8\%(20\%)&27\%(18\%)&4\%(22\%)&9\%(24\%)&2\%(12\%)&6\%(26\%)&37\%(18\%)&8\%(20\%)&29\%(19\%)&4\%(21\%)&9\%(24\%)&7\%(38\%)\\
30\%&16\%(97\%)&49\%(35\%)&5\%(20\%)&18\%(18\%)&3\%(25\%)&7\%(27\%)&2\%(16\%)&6\%(38\%)&41\%(30\%)&7\%(26\%)&27\%(27\%)&4\%(26\%)&10\%(37\%)&6\%(46\%)\\
40\%&12\%(97\%)&39\%(38\%)&7\%(37\%)&18\%(24\%)&6\%(58\%)&13\%(65\%)&5\%(56\%)&6\%(53\%)&42\%(40\%)&8\%(39\%)&26\%(34\%)&4\%(35\%)&10\%(53\%)&5\%(48\%)\\
50\%&9\%(97\%)&34\%(41\%)&8\%(51\%)&25\%(41\%)&7\%(83\%)&12\%(77\%)&5\%(61\%)&6\%(58\%)&42\%(50\%)&8\%(52\%)&29\%(48\%)&3\%(40\%)&8\%(55\%)&4\%(52\%)\\
60\%&8\%(97\%)&35\%(50\%)&9\%(74\%)&27\%(52\%)&6\%(92\%)&11\%(84\%)&5\%(73\%)&5\%(63\%)&40\%(58\%)&8\%(63\%)&30\%(59\%)&4\%(60\%)&9\%(70\%)&4\%(61\%)\\
\end{tabular}}
}
\caption{Changes of occurrence of different classes due to data selection with a DBN-based active learning approach compared to random data selection during the instrument presence task \protect\subref{tab:dis:ins} and the segment-based phase segmentation task \protect\subref{tab:dis:ins}. The values indicate the percentage of samples of each class in the training set before each annotation round and, in parenthesis, the percentage of all occurrence of a class contained in the current training set.}
\label{tab:dis}
\end{table}
\section{Conclusion}
In this paper, we presented, to the best of our knowledge, the first DBN-based active learning approach for annotating data related to surgical workflow tasks.
Our focus, in particular, was on instrument presence detection and workflow analysis.
Also we presented the first DBN-based active learning approach for video annotation.
Furthermore, we showed that our approach for selecting the next data points for annotation outperforms a random baseline and we were able to demonstrate that partially annotating videos is a valid strategy for training CNNs for surgical workflow segmentation.

Even though the results seem promising, we see potential for improving performance.
The step size of 10\% for selecting data points might not be optimal as it could encourage unnecessary redundancy in the data, as it can be assumed that similar images have a similar uncertainty.
Opting for a smaller step size might mitigate this problem.
Furthermore, incorporating a form of similarity measure in the acquisition functions might also be appropriate.
Currently we retrain our network from scratch after new labeled data has been acquired.
An alternative strategy would be to instead fine-tune the existing network, though this requires research into metrics for determining from which previous state to fine-tune.

\small{
\textbf{Conflict of interest}
S. Bodenstedt, D. Rivoir, A. Jenke, M. Wagner, M. Breucha, B. M{\"u}ller-Stich, S. Mees, J. Weitz and S. Speidel declare that they have no conflict of interest.

\textbf{Ethical approval}
For this type of study formal consent is not required.

\textbf{Informed consent}
This article contains patient data from publicly available datasets.}
\bibliographystyle{spmpsci}      
\bibliography{paper}   

\end{document}